\documentclass[11pt,a4paper]{article}

\PassOptionsToPackage{breaklinks}{hyperref}
\usepackage{xurl}

\usepackage[hyperref]{acl2021}
\usepackage{times}
\usepackage{latexsym}
\usepackage{tabularx}
\usepackage{enumerate}
\usepackage{booktabs}

\usepackage{graphicx}  

\usepackage{microtype}

\aclfinalcopy 
\def\aclpaperid{168} 

\title{ZJUKLAB at SemEval-2021 Task 4:\\ Negative Augmentation with Language Model for Reading Comprehension of Abstract Meaning}

\author{
Xin Xie\textsuperscript{\rm 1,2 \thanks{ Equal contribution and shared co-first authorship.} },  Xiangnan Chen\textsuperscript{\rm 1,2 \footnotemark[1]} , Xiang Chen\textsuperscript{\rm 1,2 \footnotemark[1]} , 
Yong Wang\textsuperscript{\rm 3}, \\ \textbf{Ningyu Zhang}\textsuperscript{\rm 1,2 \dag}, \textbf{Shumin Deng}\textsuperscript{\rm 1,2}, \textbf{Huajun Chen}\textsuperscript{\rm 1,2 \footnote{Corresponding author.}} \\
	\textsuperscript{\rm 1} Zhejiang University \& AZFT Joint Lab for Knowledge Engine \\
	\textsuperscript{\rm 2} Hangzhou Innovation Center, Zhejiang University
	\textsuperscript{\rm 3} Microsoft \\
	\texttt{\{xx2020,xnchen2020,xiang\_chen\}@zju.edu.cn} \\
	\texttt{\{zhangningyu,231sm,huajunsir\}@zju.edu.cn} \\
	\texttt{wangyon@microsoft.com}
	}

\date{}

\begin{document}
\maketitle


\begin{abstract}
This paper presents our systems for the three Subtasks of SemEval Task4: Reading Comprehension of Abstract Meaning (ReCAM). We explain the algorithms used to learn our models and the process of tuning the algorithms and selecting the best model. Inspired by the similarity of the ReCAM task and the language pre-training, we propose a simple yet effective technology, namely, negative augmentation with language model. Evaluation results demonstrate the effectiveness of our proposed approach. Our models achieve the \textbf{4th} rank on both official test sets of Subtask 1 and Subtask 2 with an accuracy of 87.9\% and an accuracy of 92.8\%, respectively\footnote{Our implementation is publicly available at \url{https://github.com/zjunlp/SemEval2021Task4}}. We further conduct comprehensive model analysis and observe interesting error cases, which may promote future researches.
\end{abstract}

\section{Introduction}
Past decades have witnessed the huge progress of representation learning in Natural Language Processing (NLP). 
With pre-trained language models, machine reading comprehension (MRC) models can extract answers from given documents and even yield better performance than humans on benchmark datasets such as Squad  \cite{rajpurkar-etal-2016-squad}. 
However, these successes sometimes lead to the hype in which these models are being described as ``understanding" language or capturing ``meaning" \cite{bender-koller-2020-climbing}. 
Note that the intention of MRC is letting the systems read a text like human beings, extracting text information and understanding the meaning of a text then answering questions, which means the systems can not only conclude the semantic of the text but also comprehend the abstract concepts under the constraint of general knowledge regarding the world \cite{DBLP:journals/corr/WangJ16a}. Nevertheless, little works as well as benchmarks focus on this direction. 

\textbf{SemEval 2021 Task4} \cite{zheng-2021-semeval-task4} is an MRC  task that focuses on evaluating the model's ability to understand abstract words.
Reading Comprehension of Abstract Meaning (ReCAM) task is divided into three Subtasks including \textbf{Subtask 1}: ReCAM-Imperceptibility, \textbf{Subtask 2}: ReCAM-Nonspecificity and \textbf{Subtask 3}: ReCAM-Intersection.
Unlike previous MRC datasets such as CNN/Daily Mail \cite{hermann_teaching_2015}, SQuAD \cite{squad}, and CoQA \cite{coqa} that request computers to predict concrete concepts, e.g. named entities. This task challenges the model's ability to fill the abstract words removed from human-written summaries based on the English context.

Note that this task's input format is similar to the MLM pre-training task of BERT \cite{devlin_bert_2019}, which aims to predict the mask tokens. 
Pre-trained language models (PLMs) such as BERT  \cite{devlin_bert_2019}, RoBERTa \cite{roberta}, ALBERT \cite{lan_albert_2020}, DeBERTa \cite{he_deberta_2021}  have achieved success on MRC tasks. 
Inspired by this, we introduce a simple yet effective method, namely, \textbf{N}egative \textbf{A}ugmentation with \textbf{L}anguage model (\textbf{NAL}) in \textbf{SemEval 2021 Task4}. 
Specifically, we augment the answer distribution with an additional negative candidate from the mask language model's prediction. 
Previous work \cite{petroni-etal-2019-language,zhou2020evaluating} indicates that the pre-trained language model has already captured much world knowledge.
Thus, we argue that knowledge can help guild the model training and identify those ambiguous abstract meanings. 
Further, we introduce other technologies such as label smoothing, domain-adaptive pre-training in our system.
We describe the detailed approaches used for the Subtasks in Section \ref{sect:system-overview}.

We conduct comprehensive experiments in Section \ref{sect:system-overview}, and we achieve the 4th system for Subtask 1: ReCAM-Imperceptibility and the 4th system for Subtask 2: ReCAM-Nonspecificity in the leaderboard. In our experiments, we observe that  PLMs without fine-tuning can easily get 60+\% accuracy on both Subtask 1 and Subtask 2, demonstrating that pre-trained language models already capture some abstract meanings. We further find that our negative augmentation with language model can improve the performance with \textbf{2.6\%} in Subtask 1 and \textbf{4.6\%} in Subtask 2. Finally, we conduct error analysis to promote future researches. 

\section{Background}
Machine reading comprehension (MRC) has received increasing attention recently, which is a challenging task.
According to the type of the answer, reading comprehension tasks can be divided into four categories \cite{chen-2018neural}: 
1) Cloze-style: The question contains a "@placeholder," and the system must choose a word or entity from the set of candidate answers to fill in the "@placeholder" to make the sentence complete.
2) Multiple choice: In this type of task, Choosing a suitable answer from K sets of given answers. This answer can be one word or a sentence. 
3) Span prediction: This kind of task is also called (Extractive question answering), which requires the system to extract a suitable range of text fragments from a given original text based on the question as to the answer. 
4) Free-form answer: This task allows the answer to be any type of text, which is necessary to mine deep-level contextual semantic information according to a given question and a collection of candidate documents, and even combine multiple articles to give the best answer. 

\begin{table}[htb]
\centering
\begin{tabularx}{0.48\textwidth}{p{0.1cm} X}
\toprule[1pt]
\textbf{P:} & Briton Davies won F42 shot put gold with a Games record at Rio 2016, but was unable to defend his 2012 discus title as it did not feature in Brazil. "I don't normally say what I'm going for," said the Welshman, 25. "But this time \textit{I'm definitely going for the two golds in both disciplines} and nothing will be better than being in front of a home crowd." ... \\
\textbf{Q:} & Paralympic champion Aled Sion Davies @placeholder two gold medals at the 2017 World Para Athletics Championships in London. \\
\textbf{A:} & (A) suffered (B) promoted (C) remains \textbf{(D) wants} (E) achieved \\
\hline

\textbf{P:} &... Low vitamin D levels can lead to brittle bones and rickets in children. The figures from the HSCNI show a dramatic rise in Vitamin D prescriptions over the last 10 years: The data does not include Vitamin D bought over the counter...\\
\textbf{Q:}&Rickets does not have the ring of a 21st Century problem - it sounds more like the @placeholder of a bygone era .\\
\textbf{A:} & \textbf{(A) horror} (B) size (C) fate (D) tale (E) death\\
\toprule[1pt]
\end{tabularx}
\caption{\label{font-table} Examples of the \textbf{SemEval 2021 Task 4}. 
Given a passage and a question, the model needs to pick the best one from the five candidates to replace @placeholder. }
\label{tb:task}
\end{table}

In \textbf{SemEval 2021 Task4}, it requires the system to have a strong ability of reading comprehension not only because the task is the cloze-style format as mentioned above but also the abstract words in answers. 
There are two definitions of abstract words: imperceptibility and nonspecificity.
Concrete words refer to things, events, and properties that we can perceive directly with our senses \cite{spreen_parameters_1966, turney_literal_2011}.
Compared to concrete words like "trees" and "red," abstract words for imperceptibility are created by humans instead of pointing the things in the natural world.
For example, as shown in Table \ref{tb:task}, "want" and "achieve" means a person's attitude towards something and a person's accomplishment about something. 
Meanwhile, the abstract words for nonspecificity can be described as upper words.
By determining whether one word can generalize another word, we can get dictionaries of different levels. 
The words with higher levels are the nonspecificity words. 
Compared to concrete concepts like groundhog and whale, hypernyms such as vertebrate are regarded as more abstract \cite{changizi_economically_2008}.

The difference between Subtask 1 and Subtask 2 is the definition of abstract words.
So the input of both Subtask 1 and Subtask 2 are the same.
The input of these tasks are shown in Table \ref{tb:task}, it can be represented as a triple $<P,Q,A>$, where $P=s_1,s_2,...,s_m$ is the passage from CNN daily  \cite{hermann_teaching_2015}, $Q$ is a human-written summary based on the passage with one abstract word replaced by "@placeholder" and $A$ is a set of candidate abstract words for filling in the "@placeholder" in the question. 

\section{System Overview}
\label{sect:system-overview}
\subsection{Model Design}

Recently, with the development of the large Pre-trained Language Models (PLMs), such as GPT \cite{gpt}, BERT  \cite{devlin_bert_2019}, RoBERTa \cite{roberta}, ALBERT \cite{lan_albert_2020}, DeBERTa \cite{he_deberta_2021}, have overwhelm the NLP community \cite{zhang2020conceptualized}.
The powerful semantic feature extraction capabilities of the PLMs make us only need to make better use of the BERT-like model itself for downstream tasks instead of adding different layers to the model. 

Similar to the normal multi-choice task, we have five candidates, one passage, and one question per sample.
Here we leverage PLMs as encoders to capture the global context representation about the passage, question, and answer.
Then a decoder is used to determine the score of each  $<P,Q,A>$ pair. 
Since we get ${A_1,...,A_n}$ $n$ answers, for every passage, we construct $n$ input samples as $[Q-A_i;P]$, the concatenation of $Q-A_i$ and $P$ .
Because the question is the summary with an abstract word removed.
We construct $Q-A$ by replacing "@placeholder" with the option from the candidate set instead of concatenating $Q$ and $A$.
After encoding all $n$ inputs for a single passage, we get the global representations $T_i$ for different options in the candidate set.
During fine-tuning PLMs,  the first special token \texttt{[CLS]} represents the global meaning of the whole input.
We use an dense decoder layer to compute the score for all $T_i$, the calculation of score is as follow:

\begin{equation}
    T_{i}= PLM( Q-A;P )
\end{equation}

\begin{equation}
 score _{i}=\frac{\exp \left(f\left(T_{i}\right)\right)}{\sum_{i^{\prime}} \exp \left(f\left(T_{i^{\prime}}\right)\right)}
\end{equation}
where the $[Q-A;P]$ is the input constructed according to the instruction of PLMs and MRC tasks, and the $T_*$ is the final hidden state of the first token \texttt{[CLS]}.
The candidate answers with higher \textit{scores} will be identified as the final prediction. 

\begin{figure}[ht]
    \centering
    \includegraphics[width=0.5\textwidth]{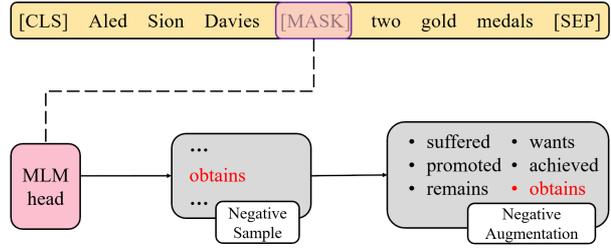}
    \caption{The procedure of Negative Augmentation with Language Model (NAL).}
    \label{fig:NAL}
\end{figure}

Since previous research  \cite{DBLP:journals/corr/abs-2012-15723,xlnet} demonstrate that there exists a gap between language model pre-training and fine-tuning the models in the downstream task and inspire by the similar task definition as MLM, we introduce the negative augmentation with language model mechanism (Section \ref{sect:approach-gap}). Note that the additional label will enhance the discriminability of the abstract meanings in a contrastive manner. In other words, the model is encouraged \textbf{NOT} to generate those abstract tokens from the language model, but the golden candidates from the given documents.  We further introduce the label smoothing (Section \ref{sec_label}), which can enhance the model performance. Finally, we leverage task-adaptive pre-training (Section \ref{sec_pretrain}) inspired by \cite{noauthor_200410964_nodate} to obtain better performance. 

\begin{figure*}

\centering
\includegraphics[width=1\textwidth]{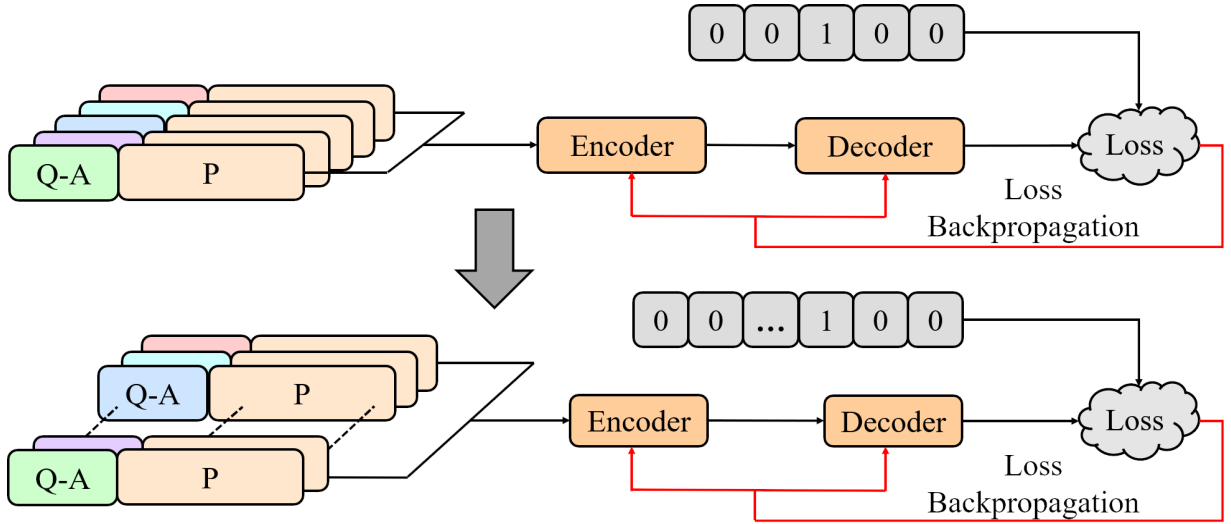}
\caption{\label{font-figue} System overview (Best viewed in color.). The top of the Figure refers to the normal fine-tuning of multi-choice models, ignoring the form of pre-training tasks. 
While the bottom of the Figure refers to our system with Negative Augmentation with Language Model (NAL), which uses the abstract words predict by the original PLM as negative candidates to augment fine-tuning.}
\label{fig: model}
\end{figure*}

\subsection{Negative Augmentation with Language Model}
\label{sect:approach-gap}

Inspired by the same format of MLM and this task, we first conduct a toy experiment to test whether a PLM can get the right answer without any supervised signal.
Firstly we replace the "@placeholder" with \texttt{[MASK]} to reconstruct the input and ask the BERT model with MLM head to predict the word token at the \texttt{[MASK]}.
Then we calculate the similarity between the word model predict and the options from the set of candidate answers.
We set the option with the highest similarity score as the model's choice.
Then we find that the BERT model without any fine-tuning gets \textbf{60+\%} accuracy in both Subtask 1 and Subtask 2.
The result above shows that PLMs have the ability to predict abstract words, and those predicted words can be leveraged as negative candidates in the fine-tuning period.

Note that huge languages have quantities of parameters; the PLMs are able to store much knowledge through pre-training tasks. 
However, \texttt{[MASK]} is not used when fine-tuning the model for downstream tasks; how to use the knowledge stored by the model on pre-training tasks more explicitly on downstream tasks has become a hot topic of current research.
Motivated by this, we try to bridge the gap between pre-train and downstream tasks.
Inspired by the contrastive learning \cite{DBLP:conf/icml/ChenK0H20,DBLP:journals/corr/abs-2010-04592} as stronger negative samples will help the model learning with better performance, we introduce our negative augmentation with language model method.
Specifically,  we let the PLMs predict the "@placeholder" replaced with \texttt{[MASK]} token to generate negative candidates.
Thus, we can leverage those negative words that may mislead the models to help train the models. Formally, we have:
\begin{equation}
P = p(m_i| \theta ,[Q-A;P]), m_i \in [1,2,...,|V|]
\end{equation}
where $P$ are the distribution of words model, predict, $m_i$ is the token in the vocabulary, and $|V|$ is the total number of the vocabulary.
We can use the distribution to get the top confusing words to augment our models, which is described in Figure \ref{fig: model}.
Due to the limitation of GPU, we add the most possible word to augment our models.

\subsection{Label Smoothing}
\label{sec_label}
Label smoothing is a well-known "trick" to improve the model's performance effectively. 
It encourages the activations of the penultimate layer to be close to the template of the correct class and equally distant to the templates of the incorrect classes \cite{muller_when_2020}. 
With more options than the original dataset by the approach mentioned in Section \ref{sect:approach-gap}, label smoothing will magnify our method's effect while fine-tuning the models.
Suppose the output of the final layer and softmax layer as follows:
\begin{equation}
p_{k}=\frac{e^{x^{T}w_{k}} }{\sum_{l=1}^{L} e^{x^{T} w_{l}}}
\end{equation}
where $p_k$ is the likelihood the model assigns to the $k$-th class, $w_k$ represents the weights and biases of the last layer.
$x$ is the vector containing the activations of the penultimate layer of a neural network concatenated with "1" to account for the bias.
let us see the equitation about the  cross entropy loss.

\begin{equation}
L = -\sum_{c=1}^{M} y_{k} \log \left(p_{k}\right)
\end{equation}

The cross-entropy formula without Label smoothing only focuses on whether the positive example is true and does not pay attention to the negative examples' relationship.
We make the soft y as follows:
\begin{equation}
y_{i}=\left\{\begin{array}{l}
(1-\varepsilon), \mbox{  right answer} \\
\frac{\varepsilon}{K-1}, \mbox{wrong answer}
\end{array}\right.
\end{equation}
We set $\varepsilon$ as $0.1$ in our models.

\subsection{Task-Adaptive Pre-training}
\label{sec_pretrain}
The BERT-like model is pre-trained in the general domain corpus such as Wikipedia. 
Since passages mainly come from CNN daily, the data distribution may be quite different from pre-training data.
Therefore, we utilize task-adaptive pre-train BERT with masked language model and next sentence prediction tasks on the domain-specific data. 
Task-adaptive pre-training not only makes the model better fit the distribution in the domain but also helps the model to predict good negative words to enhance the original dataset, which is described in Section \ref{sect:approach-gap}. 
We take two different approaches for task-adaptive pre-training as follows:

\begin{enumerate}[1)]
\item In-domain pre-training, we use the source data: CNN Daily to task-adaptive pre-training our base models\cite{sun_how_2020}. 

\item Within-task pre-training, practically we replace the "@placeholder" with the correct answer and put the same input format as the fine-tuning steps, which is $[Q-A; P]$ \cite{noauthor_200410964_nodate}.
\end{enumerate}

\section{Experimental Setup}

\subsection{Dataset}
\begin{table}[!t]
\centering
\small
    \begin{tabular}{p{4cm}cc}
         \toprule
         \textbf{Statistics / Dataset} & \textbf{Subtask 1} & \textbf{Subtask 2}  \\
         \midrule
         \# Train& 3,227& 3,318\\
         \# Trail& 1,000& 1,000 \\
         \# Dev& 837& 851\\
         \# Test& 2,025& 2,017\\
         Avg. \# Length Per Passage&262 &418   \\
         \bottomrule
    \end{tabular}
    \caption{Statistics of the \textbf{SemEval 2021 Task 4} dataset.}
    \label{dataset}
\end{table}

In Subtask 1, the training/trail/development/test contains $3,227$/$1,000$/$837$/$2,025$ instances.
In Subtask 2, the training/trail/development/test contains $3,318$/$1,000$/$851$/$2,017$ instances.
The overall statistics can be found in Table \ref{dataset}.

\subsection{Pre-processing}
For data pre-processing, we use the byte-level BPE encoding \cite{noauthor_150807909_nodate}, and the official vocabulary contains more than fifty thousand byte-level tokens. 
All tokens are stored in \textsc{merges.txt}, while \textsc{vocab.json} is a byte-to-index mapping. 
Generally speaking, the higher the frequency, the smaller the byte index.
Since the average length of the passage about Subtask 1 and Subtask 2 is $262$ and $418$, we divide those long context paragraphs.
We limit the max number of tokens in an input sample $[Q-A;P]$ to 256 for our system.
Statically, 60\% of the paragraphs exceeds the $256$ tokens (including the special tokens like \texttt{[CLS]}, \texttt{[SEP]} and so on. 
For these input samples, we divide them into new input samples with at most 256 tokens. 
To be more specific, we divide the passage to different inputs with the same question and answer.

\subsection{Hyper-parameter Setting}
Our system is implemented with PyTorch \cite{pytorch} and we use the PyTorch  version of the pre-trained language models\footnote{\url{https://github.com/huggingface/transformers} (version 3.3.0)}. 
We employ RoBERTa, ALBERT, and DeBERTa large models as our PLM encoder. 
We use AdamW optimizer \cite{loshchilov_fixing_2018} to fine-tune the models.
We set the batch size to 1,  and the max length of input to $256$ for RoBERTa, $128$ for ALBERT.

Usually, the batch size has a significant influence on the BERT-like model; due to the limit of GPU memory, we use gradient accumulation in our training steps.
We set the gradient accumulation step as $32$, which means the formal number of batch sizes is 32 in training.
We pick the best learning rate from the dev set, fine-tuning the RoBERTa, ALBERT, DeBERTa with the learning rate of $9 \times 10^{-6}$, $1 \times 10^{-5}$ and $1 \times 10^{-5}$ respectively.
We set the number of epoch to 8 for ALBERT and 12 for RoBERTa and DeBERTa.
Furthermore, we save the best model on the validation set for testing during training.
Because the formats of both Subtask 1 and Subtask 2 are the same, we set the same batch size and max length of the input sequence for training. 

\begin{table}[ht]
\centering
\begin{minipage}[t]{0.48\textwidth}
  \centering
  \makeatletter\def\@captype{table}\makeatother
  \setlength{\tabcolsep}{5mm}{
  \begin{tabular}{lccc}
    \toprule[1pt]
    \bf Model  & \bf  Dev & \bf  Test \\
    \toprule[0.5pt]
    \bf{\textit{Baseline}} \\
    RoBERTa$_{\mbox{\scriptsize Large}}$        & 83.3  & -  \\
    ALBERT$_{\mbox{\scriptsize xxLarge}}$   & 85.1  & -  \\
    DeBERTa$_{\mbox{\scriptsize Large}}$       & 84.1  & -  \\
    \bf{\textit{Ours}} \\
    RoBERTa$_{\mbox{\scriptsize Large}}$ + NAL    & 85.9  & 86.1  \\
    ALBERT$_{\mbox{\scriptsize xxLarge}}$+ NAL      & 86.2  &  85.6 \\
    DeBERTa$_{\mbox{\scriptsize Large}}$+ NAL       & 86.7  & 86.8  \\
    Ensemble                            & \bf88.5  & \bf87.9  \\
    \toprule[1pt]
  \end{tabular}}
  \caption{\label{font-table}Results (Accuracy) on  Subtask 1.}
  \label{tb:Results1}
\end{minipage}
\\[12pt]
\begin{minipage}[t]{0.48\textwidth}
  \centering
  \makeatletter\def\@captype{table}\makeatother
  \setlength{\tabcolsep}{5mm}{
  \begin{tabular}{lcc}
    \toprule[1pt]
    \bf Model & \bf    Dev & \bf  Test \\
    \toprule[0.5pt]
    \bf{\textit{Baseline}} \\
    RoBERTa$_{\mbox{\scriptsize Large}}$       & 86.7  & -  \\
    ALBERT$_{\mbox{\scriptsize xxLarge}}$  & 84.3  & -  \\
    DeBERTa$_{\mbox{\scriptsize Large}}$         & 87.7  & -  \\
    \bf{\textit{Ours}} \\
    RoBERTa$_{\mbox{\scriptsize Large}}$ + NAL    & 91.1  & 89.7  \\
    ALBERT$_{\mbox{\scriptsize xxLarge}}$+ NAL      & 89.3  &  88.6 \\
    DeBERTa$_{\mbox{\scriptsize Large}}$+ NAL       & 91.3  & 90.3  \\
    Ensemble                            & \bf93.7  & \bf92.8  \\
    \toprule[1pt]
  \end{tabular}}
  \caption{\label{font-table} Results (Accuracy) on  Subtask 2.}
  \label{tb:Results2}
\end{minipage}
\end{table}

\section{Results}

\subsection{Subtask 1 Results}
On Subtask 1 , the ReCAM-Imperceptibility task, the evaluation results are illustrated in Table \ref{tb:Results1}.
We set the three baseline models: RoBERTa$_{\mbox{\scriptsize Large}}$, DeBERTa$_{\mbox{\scriptsize Large}}$, and ALBERT${\mbox{\scriptsize xxLarge}}$.
RoBERTa$_{\mbox{\scriptsize Large}}$ + NAL, DeBERTa$_{\mbox{\scriptsize Large}}$ + NAL, and ALBERT$_{\mbox{\scriptsize Large}}$ + NAL denotes the language model with our proposed negative augmentation with language model.
Ensemble refers to the ensemble model of the three models as mentioned above with all strategies.
We find that ALBERT achieves better performance in Subtask 1 but fails to get good performance in Subtask 2, while DeBERTa and RoBERTa have better performance in Subtask 2.
Comparing with the original RoBERTa, DeBERTa, and ALBERT models, each model is hugely improved with NAL by about \textbf{2.1}\% accuracy.
We further observe that DeBERTa and RoBERTa, which have the same architecture, obtain better performance than ALBERT in the dev and test sets.
We think the possible reason is that ALBERT uses layer weight sharing, which reduces the model's generalization ability in reading comprehension, especially the abstract words meaning.
Finally, the ensemble of the best model of RoBERTa, DeBERTa, and ALBERT lead to a significant improvement (\textbf{4.3}\% accuracy) compared with baselines, which is also our final submission to the leaderboard.

\subsection{Subtask 2 Results}
On Subtask 2, the ReCAM-Nonspecificity task, the experiment results are showed in Table \ref{tb:Results2}.
Similar to the models in Subtask 1, we choose RoBERTa, DeBERTa and ALBERT as our baseline models.
All RoBERTa$_{\mbox{\scriptsize Large}}$ + NAL , ALBERT$_{\mbox{\scriptsize xxLarge}}$ + NAL and DeBERTa$_{\mbox{\scriptsize Large}}$ + NAL 
are the models with negative augmentation with language model.
Ensemble refers to the ensemble model of RoBERTa, DeBERTa, and ALBERT with all strategies.
We notice that our proposed mechanism brings significant improvement  (averaging \textbf{4.3}\% of the accuracy score) compared with baselines, demonstrating the effectiveness of our proposed strategies such as negative augmentation with a language model, label smoothing, and task-adaptive pre-training.
We observe that ensemble approach of three enhanced models (RoBERTa$_{\mbox{\scriptsize Large}}$ + NAL, ALBERT$_{\mbox{\scriptsize xxLarge}}$+ NAL and DeBERTa$_{\mbox{\scriptsize Large}}$+ NAL) obtain the best accuracy of $\bf 92.8\%$ at test set, which is also our final submit to the leaderboard.

\begin{table*}[!thbp]
\centering
\begin{tabularx}{1\textwidth}{p{3cm}  X}
\toprule[1pt]
\textbf{Example} & 
\\
\hline
\textbf{Question:} & The Aurora Borealis, better known as the Northern Lights, was spotted across @placeholder of England on Sunday.
\\
\textbf{Answer set} & \{(A) millions, \textbf{(B) parts}, (C) half, (D) isle, (E) remains\}  \\
\textbf{NAL set} & \{all, half, \textbf{parts}\} \\
\textbf{Baseline} & (C) half \\
\textbf{Model with NAL} &  (B) parts \\
\hline
\textbf{Question:} & The BBC is providing live coverage of the Scottish National Party conference in Glasgow. This live @placeholder has finished .
\\
\textbf{Answer set} & \{(A) results, (B) recording, \textbf{(C) event}, (D) action, (E) center\}  \\
\textbf{NAL set} & \{blog, recording , stream\} \\
\textbf{Baseline} & (B) recording \\
\textbf{Model with NAL} &  (C) event \\
\toprule[1pt]
\end{tabularx} 
\caption{\label{font-table} We can clearly see the negative options can help the model better understand the abstract meaning in the passage and question. Answers are \textbf{bold} in the Table. }
\label{tb:NAL-example}
\end{table*}

\subsection{Subtask3 Results}
Subtask3 focuses on the model's transferability. 
During the evaluation period, we use the data on Subtask 2 to evaluate the models trained on the Subtask 1 and vice versa.
We obtain the 82\% accuracy of the model trained on Subtask 1 and evaluated on Subtask 2 on the dev set.

During experiments for all tasks, we have tried to use different decoders like MLP and other network architecture. Eventually, we find that it does not help to improve the system's performance. 
An explanation is that the pre-trained language models (PLMs) have already captured global contextual sentence meaning at the \texttt{[CLS]} token. 

\subsection{Further Analysis}
\label{sec:error-analysis}

\subsubsection{Analysis of Negative Augmentation with Language Model}
During our experiments, we conduct case studies to figure out how our method of NAL helps the model to boost performance.
From Table \ref{tb:NAL-example}, we notice that the original PLM considers using the "all", "half" as its choice instead of "parts".
Although fine-tuned on the downstream task, the baseline model still choose "half".
In our NAL method, we add some misleading negative words to help models correct the knowledge learned from the pre-training task.

\subsubsection{Analysis of Passage Length}
\begin{figure}[h]
    \centering
    \includegraphics[width=0.5\textwidth]{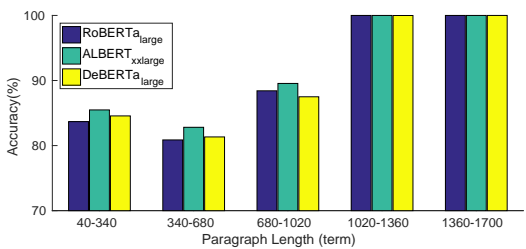}
    \caption{\label{font-table}Results (Accuracy) on  Subtask 1 with the length of passage.}
    \label{fig:result1-length}
\end{figure}

\begin{figure}[h]
    \centering
    \includegraphics[width=0.5\textwidth]{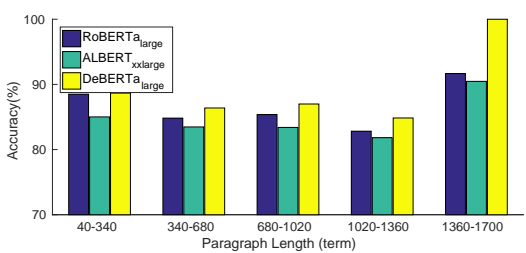}
    \caption{\label{font-table}Results (Accuracy) on  Subtask 2 with the length of passage.}
    \label{fig:result2-length}
\end{figure}

In usual MRC tasks, the length of the passage is a key factor for the models to solve the problems. We conduct experiments to analyze the performance regarding different lengths of passage. Contrary to the common assumption, from Figure \ref{fig:result1-length} and Figure \ref{fig:result2-length}, we observe that the instances with long passage obtain better performance. We think that abstract mean understanding may need comprehensive context information from the long sentence, and we will conduct further analysis in future works.

\subsubsection{Case Study}
We select four kinds of different types of error cases to promote further researches.
We classify the examples according to the main causes (pre-training, fine-tuning, and so on) of the error.
We think it will help us better understand what the model learns from pre-training and fine-tuning.
\subsubsection*{Case 1 - Influenced by the original pre-training task}
\begin{itemize}
\setlength{\itemsep}{0pt}
\setlength{\parsep}{0pt}
\setlength{\parskip}{0pt}
    \item \textbf{Passage:} "...found the United States to have the highest number of sleep deprived students, with 73\% of 9 and 10 year olds and 80\% of 13 and 14 year olds identified by their teachers as \textit{being adversely affected}. The BBC's Jane O'Brien reports."
    \item \textbf{Question:} Sleep deprivation is a significant hidden factor in lowering the @placeholder of school pupils , according to researchers carrying out international education tests .
    \item \textbf{Answer:} (A) morale (B) IQ (C) mortality (D) closure (E) achievement
    \item \textbf{Negative augmented choice:} (F) intelligence 
    \item \textbf{Right Option:} (E) achievement
    \item \textbf{Wrong Option:} (B) IQ
    \item \textbf{Potential causes:} After pre-training on the large general domain corpus, PLMs have a huge bias on predicting the \texttt{[MASK]} token. Just like the "IQ" model predict in the "@placeholder". Even after fine-tuning, our models still cannot recognize the strong evidence "being adversely affected". In our daily life, we wouldn't hold that being adversely affected by lack of sleep can lead to a decrease in IQ. We usually say that lack of sleeping may lower one's achievement in the future.
    \item\textbf{How to help models?} To prevent the model from relying too much on pre-training tasks, we create more negative samples to help the model to understand what is wrong or right about the abstract words.
\end{itemize}

\subsubsection*{Case 2 - Adverse affected by fine-tuning}
\begin{itemize}
\setlength{\itemsep}{0pt}
\setlength{\parsep}{0pt}
\setlength{\parskip}{0pt}
    \item \textbf{Passage:} " 17 May 2017 Last updated at 12:44 BST Adrien Gulfo, wearing red, who plays for the Swiss side Pully Football, tried to clear the ball away from his goal with a spectacular bicycle kick. Unfortunately for him it all went very wrong - watch the video... There was a happy ending to the story for Gulfo though, Pully went through to the cup final on penalties after the match finished 3-3."
    \item \textbf{Question:} You won't believe this own goal that was @placeholder in the Swiss lower league !
    \item \textbf{Answer:} (A) scored (B) born (C) eliminated (D) closed (E) beaten
    \item \textbf{Negative Augmented Choice:} (F) scored  (model predict "scored." Because it is the right answer, so we choose another choice "played" as an augmented choice. )
    \item \textbf{Right Option:} (A) scored
    \item \textbf{Wrong Option:} (E) beaten
    \item \textbf{Potential Causes:} It is quite weird that the original PLMs can predict the right answer, but fail to make it after fine-tuning. 
    We suppose that in the process of fine-tuning, the inconsistency of abstract vocabulary prediction and the interference of other vocabulary caused the model's effect in some cases to decrease instead.
    \item\textbf{How to help models?} We could use our approach of NAL to increase the weight of the knowledge learned in the pre-training task or leverage external knowledge \cite{zhang-etal-2019-long,DBLP:conf/www/ZhangDSCZC20,DBLP:conf/coling/YuZDYZC20,DBLP:conf/emnlp/ZhangDLCZC20}.
\end{itemize}

\subsubsection*{Case 3 - Obscure abstract word meaning}
\begin{itemize}
\setlength{\itemsep}{0pt}
\setlength{\parsep}{0pt}
\setlength{\parskip}{0pt}
    \item \textbf{Passage:} " ...Mr Habgood said: "We're pretty sure it will be popular because it was when East Street was closed for other reasons and we want to make it a friendlier place to be. "It does fit with our larger objectives to improve the town and make it safer for cyclists and pedestrians." ..."
    \item \textbf{Question:} Three busy town center streets are to be pedestrianised in a bid to improve @placeholder for shoppers and cyclists .
    \item \textbf{Answer:} (A) opportunities (B) services (C) quality (D) disruption (E) safety
    \item \textbf{Negative Augmented Choice:} (F) access 
    \item \textbf{Right Option:} (E) safety
    \item \textbf{Wrong Option:} (B) services
    \item \textbf{Potential Causes:} Due the limit of GPU memory, we cannot put the long passage into the model once a time.
    So during the training, the model can only see a small chunk of the passage, so that it cannot get the global representation of the passage.
    \item\textbf{How to help models?} We chunk those long sentences with the approach of the sliding window to help the model understanding the whole passage.
\end{itemize}

\subsubsection*{Case 4 - Hypernyms is not always right}
\begin{itemize}
\setlength{\itemsep}{0pt}
\setlength{\parsep}{0pt}
\setlength{\parskip}{0pt}
    \item \textbf{Passage:} " North Wales Fire and Rescue Service was called to Express Linen Services on Vale Road in Llandudno Junction just before 19:30 GMT on Thursday. North Wales Police said a man was treated at the scene for smoke inhalation. Police have asked people to avoid the area..."
    \item \textbf{Question:} A number of @placeholder have been evacuated as firefighters tackle a blaze at a commercial laundry firm 's premises in Conwy county.
    \item \textbf{Answer:} (A) families (B) properties (C) water (D) disruption (E) vehicles
    \item \textbf{Negative Augmented Choice:} (F) homes 
    \item \textbf{Right Option:} (B) properties
    \item \textbf{Wrong Option:} (A) families
    \item \textbf{Potential Causes:} Hypernyms is the main focus of Subtask 2, the model may consider the "families" as the upper level of the "people" occur in the passage and choose the "(A) families" instead of the right answer "(B) properties".
    \item\textbf{How to help models?} We try to use the proposed NAL to add more abstract words learned from the pre-training to mitigate this issue.
\end{itemize}

\section{Conclusion}
This paper presents our system design for the SemEval 2021 Task4.
We propose a simple yet effective method called negative augmentation with language model.  
Comprehensive experiments demonstrate the effectiveness of our proposed approach. We also conduct case studies and investigate why the model fails to obtain the correct prediction. 

Note that language models are pre-trained from the huge corpus; recently, researchers have identified the bias in the language model, which may mislead the model prediction. 
Our proposed negative augmentation with language model can help the model better discriminate candidates in fine-tuning, thus boost the performance.
From another perspective, as depicts in Section \ref{sect:approach-gap}, the language model without any fine-tuning gets \textbf{60+\%} accuracy in both Subtask 1 and Subtask 2.
This indicates that bias exists in the datasets (Part of the abstract meaning can be obtained from the language model).
More strong benchmarks should be constructed in the future. 

\section{Acknowledgments}
We  want to express gratitude to the anonymous reviewers for their hard work and kind comments. This work is funded by 2018YFB1402800/NSFC91846204/NSFCU19B2027.

\bibliographystyle{acl_natbib}
\bibliography{anthology,acl2021}


\end{document}